\documentclass[10pt,twocolumn,letterpaper]{article}

\usepackage[pagenumbers]{cvpr} 
\usepackage{graphicx}
\usepackage{amsmath}
\usepackage{amssymb}
\usepackage{booktabs}
\usepackage{breqn}
\usepackage{graphicx}
\usepackage{booktabs}
\usepackage{tabularray}
\usepackage{rotating}
\usepackage{amssymb}
\usepackage{subcaption}
\usepackage{adjustbox}
\usepackage[dvipsnames]{xcolor}
\definecolor{cvprblue}{rgb}{0.21,0.49,0.74}
\usepackage[pagebackref,breaklinks,colorlinks,allcolors=cvprblue]{hyperref}
\usepackage[capitalize]{cleveref}
\crefname{section}{Sec.}{Secs.}
\Crefname{section}{Section}{Sections}
\Crefname{table}{Table}{Tables}
\crefname{table}{Tab.}{Tabs.}

\usepackage[normalem]{ulem}
\usepackage{multicol}
\useunder{\uline}{\ul}{}

\def\paperID{17309} 
\def\confName{CVPR}
\def\confYear{2025}

\definecolor{purple}{rgb}{0.420, 0.204, 0.922}
\definecolor{csr}{RGB}{15, 255, 67}
\definecolor{cbm}{RGB}{77, 150, 243}
\definecolor{ppnet}{RGB}{150, 0, 250}
\definecolor{tree}{RGB}{246,0,140}
\definecolor{pip}{RGB}{230, 209, 19}
\title{Interactive Medical Image Analysis with Concept-based Similarity Reasoning}

\author{
Ta Duc Huy$^\dagger$
\and Sen Kim Tran
\and Phan Nguyen
\and Nguyen Hoang Tran
\and Tran Bao Sam 
\and Anton van den Hengel$^\dagger$
\and Zhibin Liao$^\dagger$
\and Johan W. Verjans$^\dagger$
\and Minh-Son To$^\dagger$ $^\ddagger$
\and Vu Minh Hieu Phan $^\dagger$\thanks{Corresponding author.}
\\ \small{$^\dagger$ Australian Institute for Machine Learning, The University of Adelaide,
$^\ddagger$ Flinders University}
}

\begin{document}
\maketitle


\begin{abstract}
    The ability to interpret and intervene model decisions is important for the adoption of computer-aided diagnosis methods in clinical workflows.
    Recent concept-based methods link the model predictions with interpretable concepts and modify their activation scores to interact with the model. However, these concepts are at the image level, which hinders the model from pinpointing the exact patches the concepts are activated.
    Alternatively, prototype-based methods learn representations from training image patches and compare these with test image patches, using the similarity scores for final class prediction. However, interpreting the underlying concepts of these patches can be challenging and often necessitates post-hoc guesswork.
   To address this issue, this paper introduces the novel Concept-based Similarity Reasoning network (CSR), which offers (i) patch-level prototype with intrinsic concept interpretation, and (ii) spatial interactivity. First, the proposed CSR provides localized explanation by grounding prototypes of each concept on image regions.
    Second, our model introduces novel spatial-level interaction, allowing doctors to engage directly with specific image areas, making it an intuitive and transparent tool for medical imaging. CSR improves upon prior state-of-the-art interpretable methods by up to 4.5\% across three biomedical datasets. Our code is released at \href{https://github.com/tadeephuy/InteractCSR}{https://github.com/tadeephuy/InteractCSR}. 
\end{abstract}   
\section{Introduction}
\label{sec:intro}


\begin{figure*}[t]
\centering
    \includegraphics[width=\textwidth]{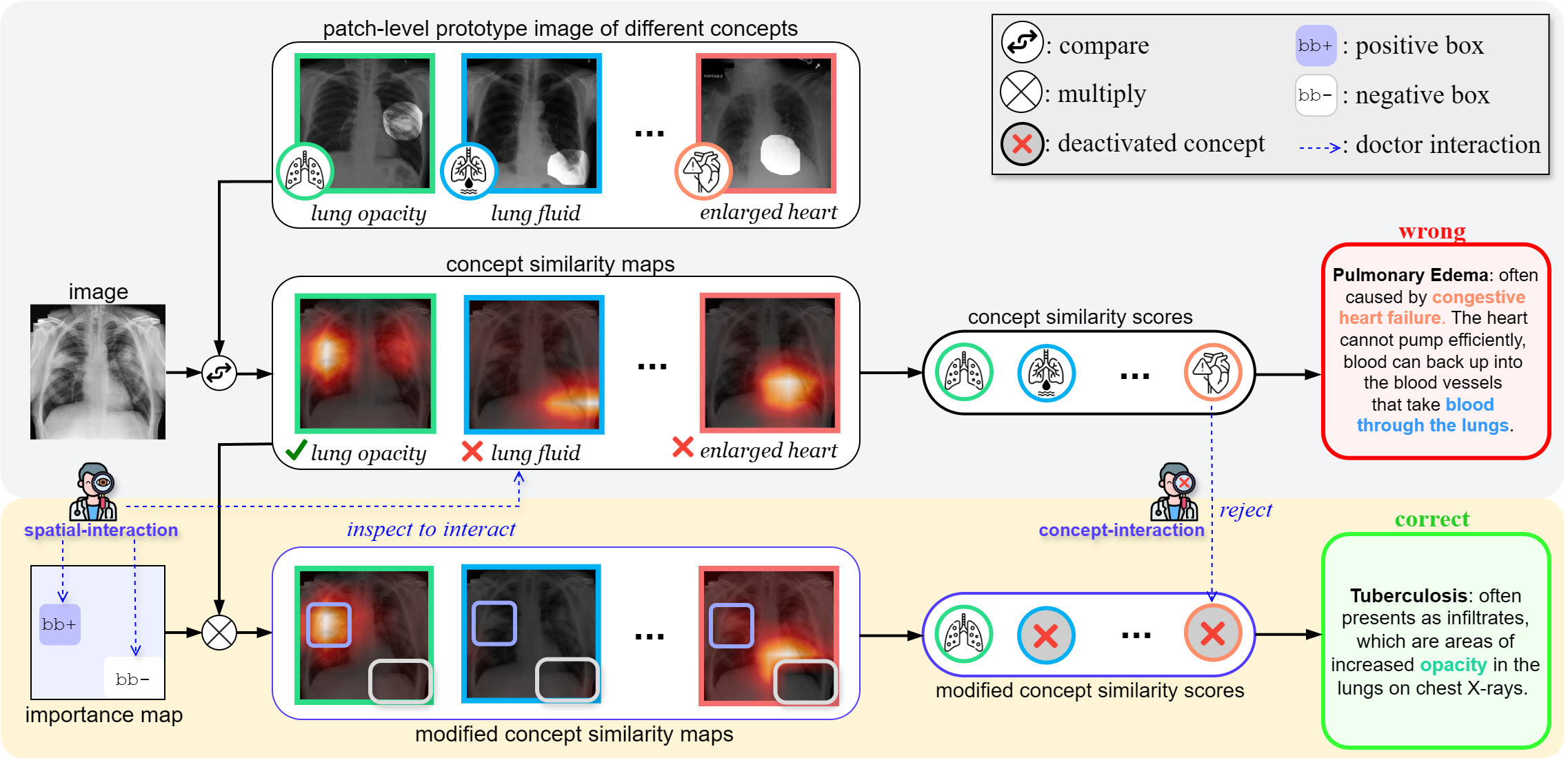}
    \caption{\textbf{Enhancing Transparency in Diagnosis Workflow with Doctor-in-the-Loop using CSR.} \textit{Top section}: 
    CSR predicts \textbf{Pulmonary Edema}, falsely associating with the two concepts \textit{lung fluid} and \textit{enlarged heart}.
    For each concept, CSR explains its prediction by comparing the input  with the highlighted region on the prototype image to create the corresponding similarity map.
    \textit{Bottom section}: As the doctor inspects the similarity maps of each concept, he suppresses incorrect \textit{lung fluid} attention, and reinforces the attention on \textit{opacity} on the left of the image through \textcolor{purple}{spatial-interaction}: ``drawing positive and negative boxes" to create an importance map indicating where to focus and ignore. Secondly, noting the heart is normal, he ``rejects" the \textit{enlarged heart} concept via \textcolor{purple}{concept-interaction}. CSR then recalibrates its prediction to \textbf{Tuberculosis}, aligning with the observed \textit{opacity}.}
    \label{fig:spatial_inter}
\end{figure*}

Despite the success of deep neural networks in medical image analysis \cite{lu2024visual,javed2024cplip,huy2022adversarial,quan2021xpgan,phan2024decomposing,Sam_2024_ACCV}, their ``black-box" nature limits their adoption in safety-critical clinical settings.
Attribution-based methods \cite{selvaraju2017grad, smilkov2017smoothgrad, zhou2016learning,wang2020score,chowdhury2024cape} 
generate saliency maps to identify influential image regions.
However, recent studies raised concerns about the reliability and accuracy of visual interpretation, potentially compromising decision-making \cite{rudin2019stop,nguyen2021effectiveness}.
To address this, Concept Bottleneck Models (CBM) predict the interpretable concepts first and use them to predict the label ~\cite{koh2020concept,kim2023concept, yang2023language,oikarinen2023label}.
CBM allows human intervention in the predicted concepts, which is shown to significantly benefit performance.
Alternatively, prototype-based methods are proposed to learn visual part prototypes to explain the classification decision. 
Specifically, ProtoPNet \cite{NEURIPS2019_adf7ee2d} and its derivatives \cite{rymarczyk2020protopshare, nauta2021neural, rymarczyk2022interpretable, Bontempelli2022ConceptlevelDO, nauta2023pip, gerstenberger2023but, wang2024mcpnet}
learn prototypical parts from training data and compare test images with these parts to compute similarity scores for final predictions.

Despite their inherent interpretability, concept-based methods only offer image-level explanations by linking images to global concepts, without identifying specific local patches correlated with these concepts.
In contrast, prototype-based methods deliver patch-level explanations but necessitate \textit{post-hoc} analysis to correlate visual patches with semantically meaningful concepts.
This challenge intensifies in medical imaging, where subtle visual medical findings make concept assignments particularly difficult. How to create concept-based models that enable patch-level explanation for medical imaging is an open-ended question.

In radiological imaging, radiologists routinely reference an atlas 
of visual examples, depicting distinctive concepts of a pathology, such as opacity, fluid presence in the lung, enlarged heart shadow 
to aid diagnosis.
Inspired by this observation, we propose a novel Concept-based Similarity Reasoning network, which offers two unique contributions. \textit{First}, our CSR gives patch-level and interpretable-by-design concepts \textit{without} requiring post-hoc analysis, in contrast to prototype-based methods~\cite{NEURIPS2019_adf7ee2d,nauta2023pip,rymarczyk2022interpretable}. We classify a disease by searching locally for the presence of visual exemplars of its concepts on the image.
For example, a cancer diagnosis on chest X-ray could involve a radiologist comparing a suspected tumor with other known malignant tumors to assess its severity ~\cite{holt2005medical}.
The presence of the tumor is a \textit{concept} to predict the target cancer.
Our framework trains a model to search locally for the presence of similar tumor exemplars (or prototypes) on every patch of the image.
After training, our model explicitly learns interpretable concept-grounded patch
prototypes without a post-hoc analysis.

\textit{Second}, our model enables spatial interactivity on the patch level.
Unlike previous methods which only intervene in model predictions on the concepts-level outputs ~\cite{Bontempelli2022ConceptlevelDO,gerstenberger2023but,yuksekgonul2023posthoc,yan2023towards}, we introduce spatial interactivity on the input image. 
Ours offers intuitive benefits in medical imaging, where precise image analysis is crucial for clinicians.
Deep learning models frequently
capture spurious shortcuts correlations rather than the intended signals—a phenomenon known as the Clever-Hans effect ~\cite{lapuschkin2019unmasking}
.
Therefore, 
integrating doctor feedback to guide the model on relevant cues significantly enhances trustworthiness in clinical settings.
Our contributions:
\begin{itemize}
    \item We propose a novel Concept-based Similarity Reasoning network (CSR) that provides concept-level explanation. To the best of our knowledge, this is the first \textit{interpretable-by-design} method that offers localized concept explainability for medical images without post-hoc analysis like part-prototype approaches. 
    \item We introduce a novel spatial interaction mechanism (\textit{where to look}), alongside train-time interaction (\textit{what not to learn}) and concept-level interaction (\textit{what not present}). CSR facilitates a comprehensive interactive framework that actively involves doctors in both train and test phases.
    \item We conducted extensive experiments on three datasets and showed that CSR achieves a significant improvement by up to 4.5\% F1-score compared to other interpretable methods.
\end{itemize}

\begin{figure*}[t]
\centering
    \includegraphics[width=0.9\textwidth]{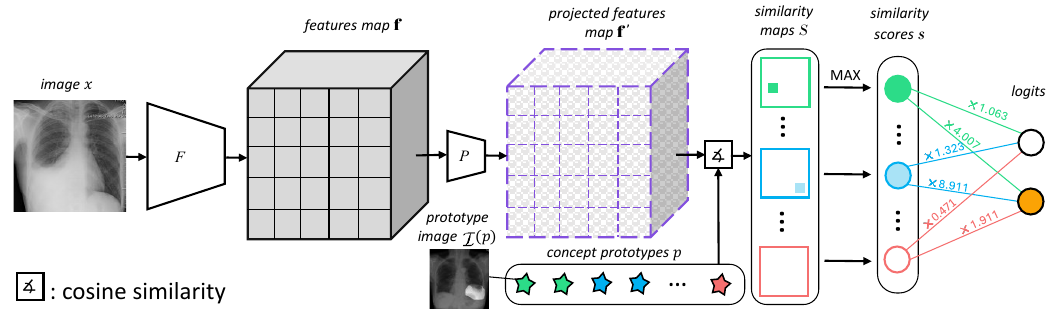}
    \centering
    \caption{Inference logic of CSR. Each concept prototype represents a specific concept from a training image. CSR generates 2D similarity maps by computing the cosine similarity between these concept prototypes and the feature maps, and considers the maximum values as similarity scores to calculate the prediction logits. The \textit{prototype image} refers to the training image associated with a specific concept prototype as  detailed in Sec.~\ref{subsec:model_exp}}
    \label{fig:inference}
\end{figure*}

\section{Method}
CSR comprises (i) a Concept model to extract interpretable concept features; (ii) a feature projector $P$ to enhance the concept feature space; and (iii) a task head $H$ to classify based on concept similarity scores.
This section 
describes the inference logic of the proposed Concept-based Similarity Reasoning model and the prototypes learning framework. Fig.~\ref{fig:inference} illustrates the inference process.
\subsection{Concept-based Similarity Reasoning}
\label{sec:sim_re}
\noindent\textbf{Concept-based classification.} 
From an input $x$, concept-based methods first predict the scores of $K$ explainable concepts $\{c^k\}_{k=1}^K$ and then use them as support to predict the target $y$.
Instead of predicting the concept scores, CSR calculates the similarity scores between $x$ and the examples (or prototypes) of each concept.
It then uses the concept similarity scores to support the prediction of $y$.

\noindent\textbf{Similarity reasoning.} 
To construct the similarity reasoning inference logic,
we establish an \textit{atlas}: a set of $M\cdot K$ concept prototypes $\{p^{k_m}\}_{k=1,m=1}^{K,M}$, where we have $M$ prototypes for each concept $c^k$.
In essence, $p^{k_m} \in \mathbb{R}^{\mathcal{C}}$ represents the $m^{\text{th}}$ example of concept $k$.
Given an input image $x$, we use the feature extractor $F$ to create a feature map $\mathbf{f} \in \mathbb{R}^{\mathcal{C}\times\mathcal{H}\times\mathcal{W}}$ having $\mathcal{H}\cdot\mathcal{W}$ feature patches.
The similarity score $s^{k_m}$ is defined as the maximum cosine similarity between the prototype $p^{k_m}$ and each feature patch $\mathbf{f}(h,w)$, quantifying the presence of $p^{k_m}$ on $x$.
As such, we can aggregate the presence of $M \cdot K$ concept prototypes $\{p^{k_m}\}$ into a similarity scores vector:
\begin{align}
    \label{eqn:sim_vector}
    s = [s^{k_m}]_{k=1,m=1}^{K,M} = [s^{1_1}, s^{1_2}, \ldots, s^{{K}_{M-1}}, s^{K_M}].
\end{align}
Then, the model predicts the target $y$ from the similarity vector $s$ with a task head $H$: $y = H(s)$.
For example, the input chest X-ray is similar to several images from the atlas that have: fluid build-up in the left lung (score 0.85), and the abnormal heart size (score 0.88). Based on these similarities, the model predicts target class \textbf{Pulmonary Edema}.
In short, the model classifies $y$ by comparing exemplars $p^{k_m}$ of the interpretable concept $c^{k}$ on each local patch on the image $x$. 
The choice of $H$ should be but not be limited to a whitebox model such as a single linear layer or a shallow tree so that the reasoning from $s \rightarrow y$ maintains interpretability.


\noindent \textbf{Discussion.} Prototype-based methods~\cite{NEURIPS2019_adf7ee2d,nauta2023pip,rymarczyk2020protopshare,rymarczyk2022interpretable} also learn a set of exemplar prototypes $p$. However, they require post-hoc analysis by experts to associate the \textit{interpretable} concept labels to the learned prototypes, which is intractable in the fine-grained medical image analysis. In stark contrast, the proposed model learns patch-level and interpretable concept prototypes \textit{by design}.

\begin{figure*}[t]
\centering
\includegraphics[width=0.9\textwidth]{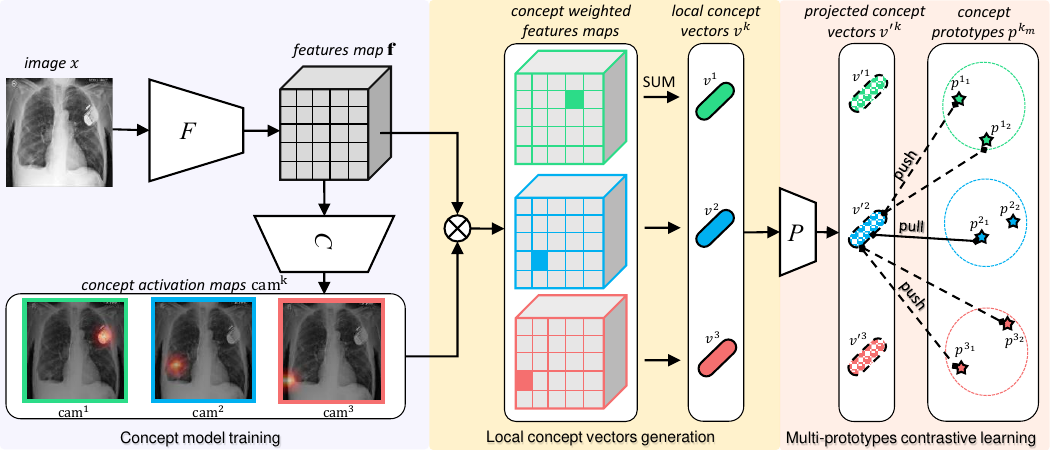}
    \caption{\textbf{The novel Concept prototypes learning framework.}
We pretrain a Concept model to generate concept activation maps.
The local concept vectors  are then generated by weighting and summing the feature maps with the activation map of the corresponding concept.
To enhance the compactness and the generalizability of the concept feature space, we introduce a novel multi-prototype learning objective. After projecting concept vectors via a projector $P$, the proposed objective pulls concept features to its nearest concept prototypes, while pushing away from prototypes of other concepts.
    }
    \label{fig:train_contrastive}
\end{figure*}

\subsection{Concept prototypes learning}
\label{sec:concept_mem}
This section describes the proposed methods to learn the concept prototypes. Overall, our method consists of a Concept model to learn interpretable concept vectors, and a novel contrastive learning objective to enhance generalizability of concept vectors. The overview of our learning framework is shown in Fig.~\ref{fig:train_contrastive}.

\noindent \textbf{Concept model.} 
The Concept model comprises the feature extractor $F$,
and the concept head $C$ that maps $\mathbf{f}$ into the one hot concept vector $c\in\mathbb{R}^K$ such that $c=C(\mathbf{f})$, indicating the presence of concepts being associated with $x$.
The concept head $C$ includes a $1\times1$ convolution layer that maps $\mathbf{f}$  to $K$ channels: $\mathbb{R}^{\mathcal{C}\times\mathcal{H}\times\mathcal{W}} \rightarrow \mathbb{R}^{K\times\mathcal{H}\times\mathcal{W}}$, representing the class activation maps $\{\text{cam}^k\in\mathbb{R}^{\mathcal{H}\times\mathcal{W}}\}_{k=1}^K$ for $K$ concepts.

\noindent \textbf{Local concept vectors acquisition.} 
If the region $(h,w)$ on $x$ contains the visual characteristics of concept $k$,
we consider the \textit{patch} feature $\mathbf{f}(h,w)$ important and select it as the local concept vector.
We first normalize  $\text{cam}^k$ using spatial softmax. 
Then, we soft-select the local concept vector $v^k \in \mathbb{R}^{\mathcal{C}}$ for concept $c^k$ by weighting $\mathbf{f}$ with $\text{cam}^k$ and summing the products spatially:
\begin{align}
    \label{eqn:softmax}
    v^k = \sum^{\mathcal{H},\mathcal{W}} \underset{h,w}{\text{softmax}}(\text{cam}^k)\cdot \mathbf{f}.
\end{align}
With a training set of $\{x_i\}^N$ images of $K$ concepts, we generate a set $\{v_i^k\}^{V\leq N\cdot K}$ of local concept vectors.

\noindent \textbf{Similarity maps as explanations.} 
Based on the local concept vector, we construct a \text{2D} similarity map for concept $c^k$ on $x$ by evaluating the cosine similarity 
between each patch feature $\mathbf{f}(h,w)$ and vector $v^k$: 
\begin{align}
    \label{eqn:cosine}
    S^k(h,w)= \cos(v^k, \mathbf{f}(h,w)) = \langle v^k, \mathbf{f}(h,w) \rangle,
\end{align}
where $\langle , \rangle$ is the dot product of the two normalized vectors. 
As $v^k$ is the importance weighted sum of $\mathbf{f}$ with respect to $c^k$, comparing it with the important patch $\mathbf{f}(h,w)$ results in a high similarity score at $S^k(h,w)$, and vice versa.
Effectively,  $S^k\in\mathbb{R}^{\mathcal{H}\times\mathcal{W}}$ is the local explanation for concept $c^k$ on $x$
.
The similarity score of $x$ with the concept $c^k$ is defined as the maximum of the similarity map $S^k$:
\begin{align}
    \label{eqn:gmp}
    s^k = \max\limits_{h,w}S^k(h,w).
\end{align}

\begin{figure}[!h]
\centering
\begin{subfigure}[b]{0.4\textwidth}
    \centering
    \includegraphics[width=\textwidth]{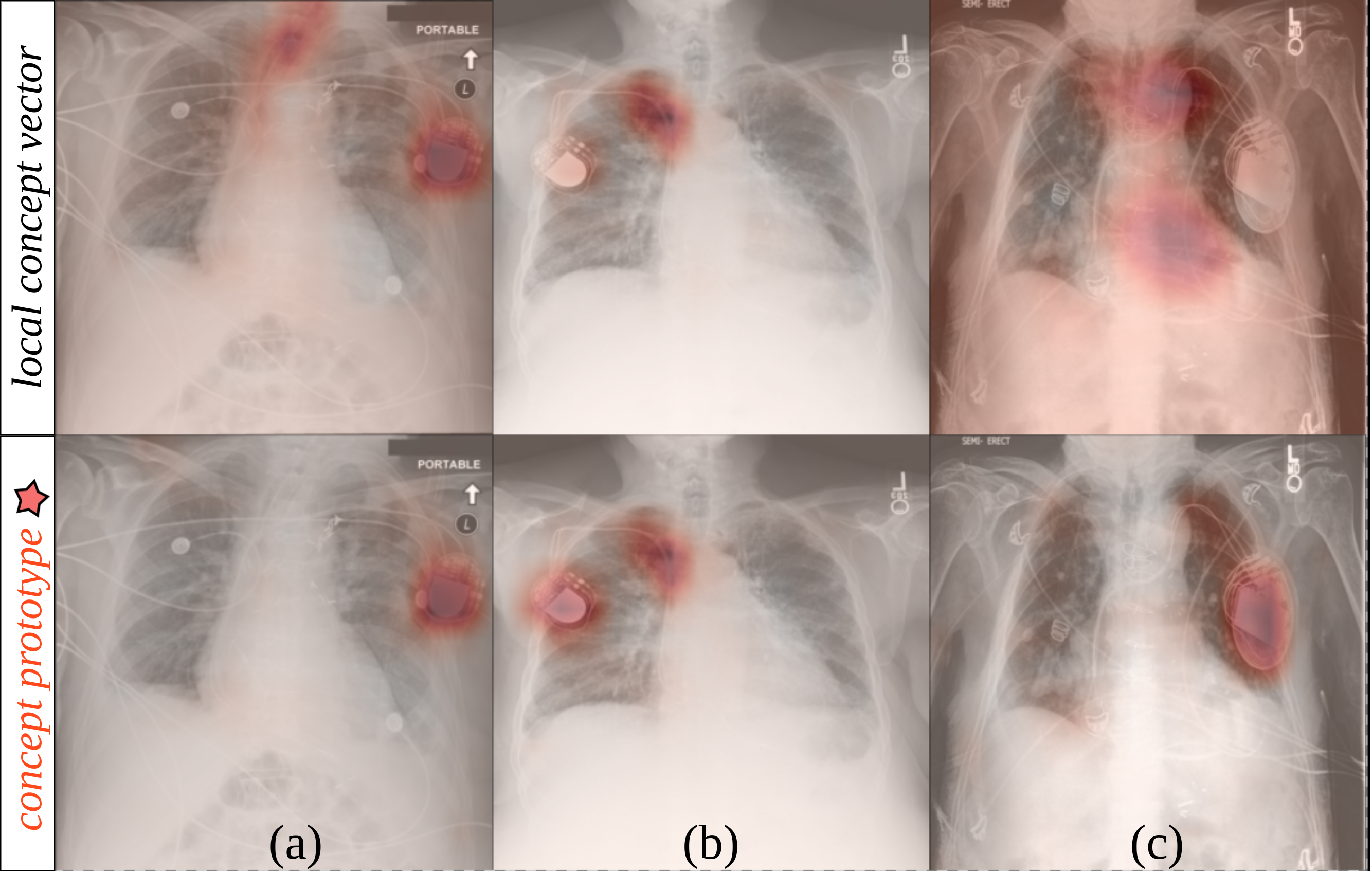}
    \vspace{1pt}
    \caption{Similarity maps generated by the local concept vector (first row) and \textcolor{red}{concept prototype} (second row) on the same image (a) and new images (b), (c). The local concept vector fails to highlight the concept \texttt{pacemaker} on new images while the concept prototype generalizes well with better localization.}
    \label{fig:sim_map}
\end{subfigure}
\hfill
\begin{subfigure}[b]{0.4\textwidth}
    \centering
    \includegraphics[width=\textwidth]{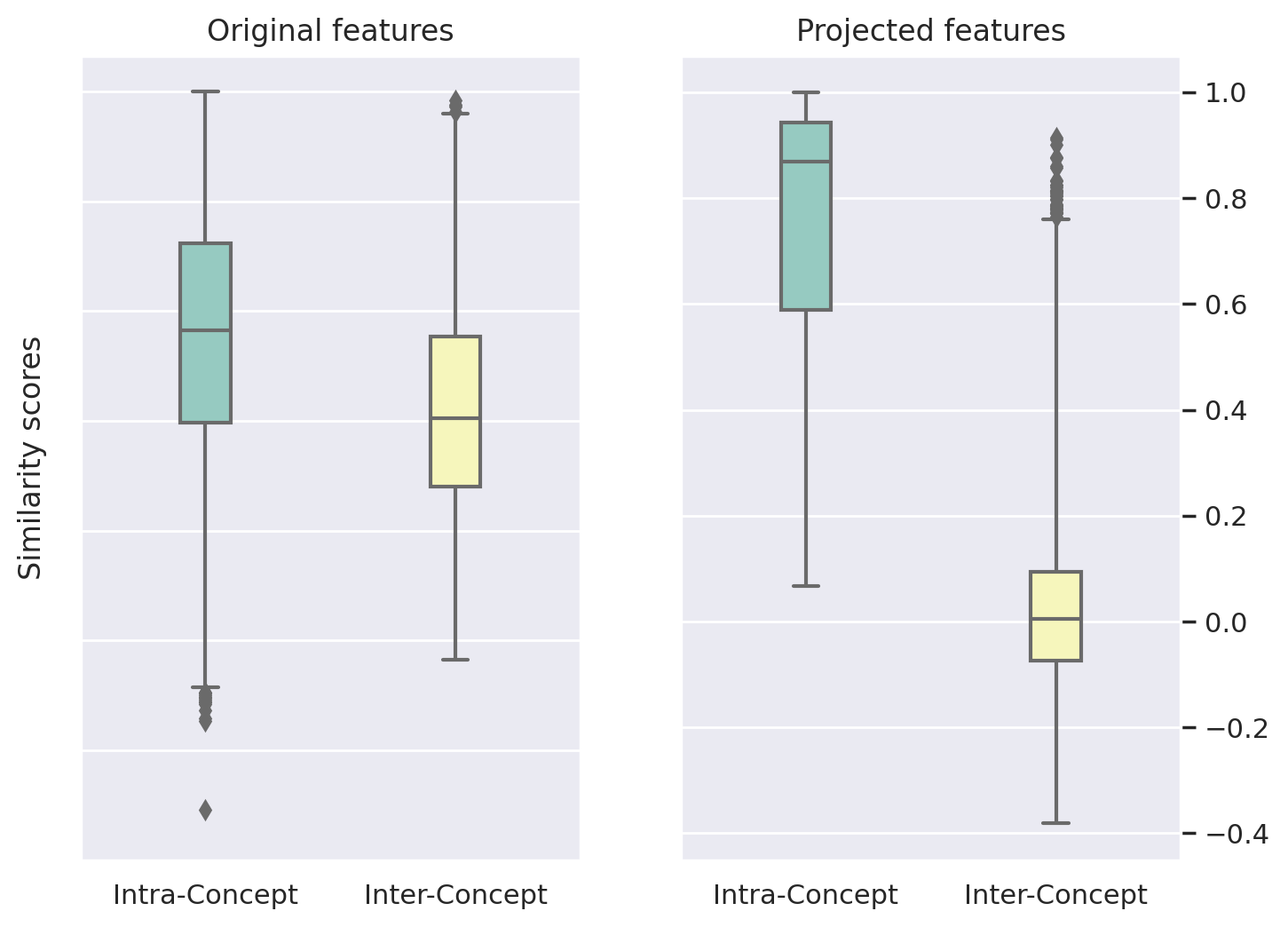}
    \caption{Comparison between the original $v$ and projected feature space $v'$. The box plots of the similarity score distribution between $v_i^k$ and $v_{j\neq i}^{k'}$. Intra-Concept: $k=k'$, Inter-Concept: $k\neq k'$. The projected feature space is more compact and has better inter-concept separation.}
    \label{fig:box_plots}
\end{subfigure}
\caption{Contrastive learning improves the concept feature space. (a) Qualitative and (b) quantitative comparison.}
\label{fig:compare}
\end{figure}

After obtaining the similarity scores $s_i^k$ for the local concept vectors $\{v_i^k\}$ with the input $x$ , one can naively adopt the inference logic in Sec.~\ref{sec:sim_re}.
However, this leads to two problems.
First, the computation complexity is $O(N \cdot K)$.
Second, $v_i^k$ is \textit{local}, which is specific to each image, thus non-generalizable to new images.
We investigate the feasibility of this approach by comparing $v_i^k$ with new images $v_{j\neq i}^{k'}$ and analyse the similarity scores distribution of intra-concept $(k=k')$ and inter-concept $(k\neq k')$.
It is observed that the concepts are not well-represented in new images.
In other words, a local concept vector $v_i^k$ generated from an image $x_i$ fails to highlight the concept $k$ on a new image $x_{j \neq i}$, as shown in Fig.~\ref{fig:compare}.

\noindent \textbf{Concept prototypes learning.}
To improve the generalizability of concept representation $v$ and reduce the number of concept vectors, 
our solution involves: (i) employing contrastive learning to enhance the shared embedding space between $v$ and $\mathbf{f}$; (ii) downsampling the concept-specific-and-sample-specific set $\{v_i^k\}$ to a small core set of  \textit{prototypes} $\{p^k\}$ that is only specific to each concept, for efficiency.

We use a projector $P$ to map $v$ to a feature space with the goal to improve inter-concept separation and intra-concept compactness such that $v'=P(v)$ via contrastive learning.
Then, we mine prototypes $p^k$ for each concept, which can be a single prototype or a diverse set of prototypes.

\textit{Single prototype per concept.}
We initialize a prototype vector $p^k\in\mathbb{R}^{\mathcal{C}}$ for each concept $c^k$.
\cite{qian2019softtriple} shows that the cross-entropy of sharpened softmax logits is equivalent to a smoothed Triplet Loss \cite{dong2018triplet} in a unit sphere.
By minimizing the maximal entropy, we can pull ${v'}_i^{\tilde{k}}$ to their respective prototype $p^{\tilde{k}}$ and push it away from irrelevant prototypes $k \neq \tilde{k}$. 
Accordingly, we learn the prototype vector $p^{\tilde{k}}$ by minimizing the term:
\begin{align}
    \label{eqn:ce_single}
    \ell_{\text{con}}({v'}_i^{\tilde{k}}) = -\log \left( \frac{\exp(\lambda \langle p^{\tilde{k}}, {v'}_i^{\tilde{k}}\rangle)}{\sum_{k \in K} \exp(\lambda \langle p^k, {v'}_i^{\tilde{k}}\rangle)} \right),
\end{align}
where $\lambda > 1$ is the scaling factor to sharpen the softmax logits, $p$ and $v'$ are $L_2$ normalized.

\textit{Multiple prototypes per concept.}
Due to the multi-modal  nature of data where one class contains several clusters rather than just a single one, we initialize  $M$ prototypes per concept $\{p^{k_m}\}_{m=1}^{M}$ such that $M\ll N$.

The similarity of ${v'}_i^{\tilde{k}}$ with its concept $c^{\tilde{k}}$ is defined as the similarity to the nearest $p^{\tilde{k}_m}$,
and can be approximated by summing the compatibility with the $M$ clusters prototypes $p^{\tilde{k}_m}$.
Specifically, we first compute the probabilistic assignment $\text{q}^{\tilde{k}}_m({v'}_i^k)$ to the prototypes:
\begin{align}
    \label{eqn:m_dis}
    \text{q}_m({v'}_i^{\tilde{k}})= \underset{m}{\text{softmax}}(\gamma\langle p^{\tilde{k}_m}, {v'}_i^{\tilde{k}}\rangle)
\end{align}
where  $\text{q}_m$ is the assignment distribution of the $M$ prototypes in $c^{\tilde{k}}$ and $\gamma>1$ is the scaling factor.
The similarity of the concept vector with each prototype is obtained by weighing their cosine similarity with the assignment:
\begin{align}
\text{sim}^{\tilde{k}_m}({v'}_i^{\tilde{k}}) = \text{q}_m({v'}_i^{\tilde{k}})\cdot \langle p^{\tilde{k}_m}, {v'}_i^{\tilde{k}}\rangle.
\end{align}
After that, we can obtain the similarity with concept $c^{\tilde{k}}$ by summing the compatibility with each cluster prototype:
\begin{equation}
\text{sim}^{\tilde{k}}({v'}_i^{\tilde{k}}) =  \sum_m^M \text{sim}^{\tilde{k}_m}({v'}_i^{\tilde{k}}).
\end{equation}
Finally, we compute the contrastive loss by maximizing the similarity of a concept vector with the cluster prototypes of the positive concept, similar to Eqn.~\ref{eqn:ce_single}:
\begin{align}
    \label{eqn:ce_multi}
    \ell_{\text{con-m}}({v'}_i^{\tilde{k}}) &= -\log \left( \frac{\exp(\lambda (\text{sim}^{\tilde{k}}({v'}_i^{\tilde{k}}) + \delta))}{\sum_{k \in K} \exp(\lambda \, \text{sim}^k({v'}_i^{\tilde{k}}))} \right), \\
    \text{if} \quad k &= \tilde{k}, \quad \text{sim}^k({v'}_i^{\tilde{k}}) = \text{sim}^k({v'}_i^{\tilde{k}}) + \delta. \nonumber
\end{align}
A small margin $\delta$ between concept prototypes is added to further expand the decision boundaries.
As a result, we learn a set of concept prototypes $\{p^{k_m}\}^{M\cdot K}$ that is significantly smaller in numbers compared to $\{v_i^k\}^{N \cdot K}$, and the projector $P$ that enhances the inter-concept separation and intra-concept compactness.
We illustrate the learning process of $P$ and $\{p\}$ in Fig.~\ref{fig:train_contrastive}.

From Eqn.~\ref{eqn:softmax} , $v_i^k$ is a soft-selected feature patch of concept $c^k$ from the original feature $\mathbf{f}_i$,
we can directly use $P$ to map feature patch $\mathbf{f}_i(h,w)$  to the compact and discriminative prototypical feature space.
We then obtain the similarity of the concept prototypes with $x$ from Eqn.~\ref{eqn:cosine}:
\begin{align}
    s^{k_m} =\max\limits_{h,w}  S^{k_m}(h,w) =\max\limits_{h,w} \langle p^{k_m}, P(\mathbf{f}(h,w)) \rangle,
\end{align}
which are aggregated to create the similarity vector $[s^{k_m}]$ in Eqn.~\ref{eqn:sim_vector}.
We quantitatively and qualitatively present the improved representation of $p$ in Fig.~\ref{fig:compare}.

    

\noindent \textbf{Training details.} Initially, we train the Concept model for multi-label classification using binary cross-entropy, predicting one-hot encoded vector of ground truth concepts.
Then, we obtain the local concept vectors and learn the concept prototypes and the projection head with a contrastive loss $\ell_{\text{con-m}}$  to enhance concept generalization.
Following, the task head is trained to predict the target from similarity scores with the concept prototypes, using cross-entropy.


\section{Doctor-in-the-loop interactivity}
\label{sec:human_inter}

\subsection{Model explanations}
\label{subsec:model_exp}
Given an input image $x$, the model returns output $\hat{y}$ and the explanations for its prediction that include:
\begin{itemize}
    \item The similarity maps $[S^{k_m}]$ to show the activated region of the concept prototype $p^{k_m}$ on $x$.
    \item The corresponding prototype image $\mathcal{I}(p^{k_m})$ as a reference, also highlighting the activated regions of $p^{k_m}$. 
    \item The concept similarity scores $[s^{k_m}]$, indicating the similarity of the two activated regions of $p^{k_m}$ on the input image $x$ and on the reference image $\mathcal{I}(p^{k_m})$.
\end{itemize}
The model explanations correspond to the similarity-based inference logic outlined in Sec.~\ref{sec:sim_re}. 
The operator $\mathcal{I}$ retrieves the image associated with the prototype $p^{k_m}$.
We replace each $p^{k_m}$ with the nearest projected local concept vector ${v'}_i^k$ and link it to its source image $x_i$: 
\begin{equation}
    \mathcal{I}(p^{k_m}) = x \leftarrow \underset{i}{\text{argmin}} \,\langle {v'}_i^{k}, p^{k_m}\rangle.
\end{equation}

\subsection{Train-time interaction}
\label{subsec:train_inter}
\begin{figure}
\centering
\includegraphics[width=0.9\columnwidth]{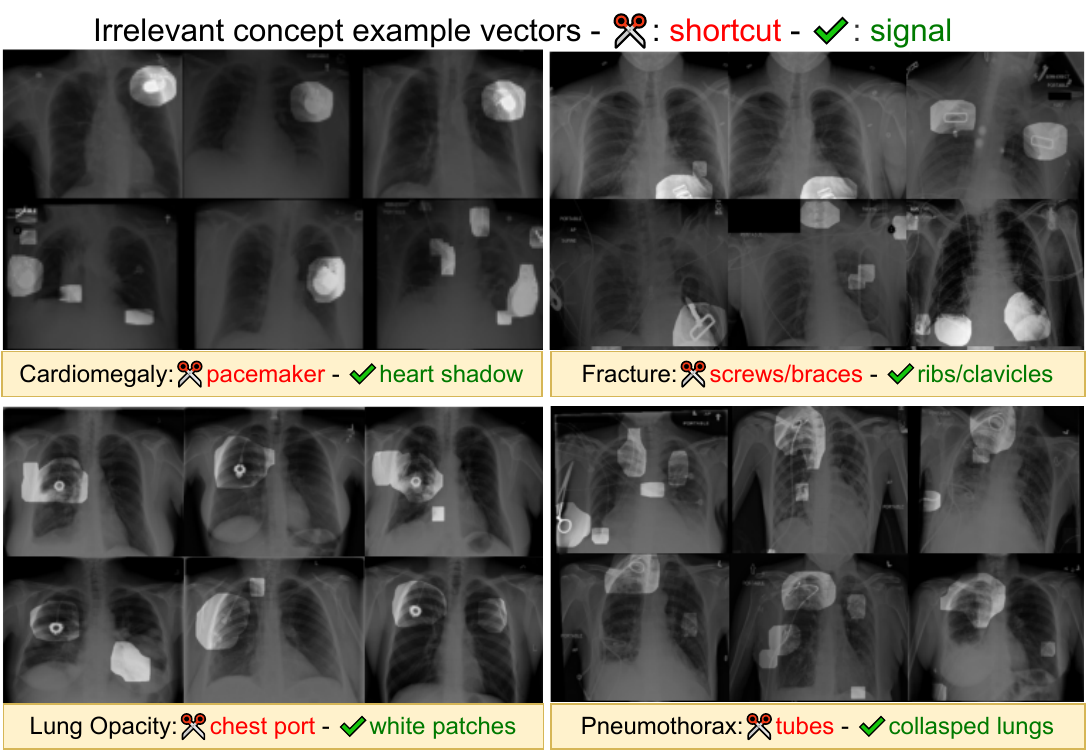}
    \caption{The irrelevant concept prototypes to be discarded in train-time interaction, highlighted at the activated regions.}
    \label{fig:shortcuts}
\end{figure}

After we learn the concept prototypes $\{p^{k_m}\}$ and its associated human-interpretable version $\{\mathcal{I}(p^{k_m})\}$, or \textit{atlas}, doctors can inspect each concept prototype and \textit{discard} any unqualified examples.
This process can reveal and mitigate the Clever-Hans effect in the Concept model by examining the clues it is taking to predict the concepts.
If a given prototype is shown to be a shortcut rather than a meaningful signal, it is removed from $\{p^{k_m}\}$ and is not used to calculate the similarity scores $[s^{k_m}]$ for the prediction of $y$, effectively nullify the Clever-Hans effect.
We illustrate the irrelevant examples of different concepts in Fig.~\ref{fig:shortcuts}.
The similarity-based inference logic allows this interaction as we only need as much as the remaining relevant examples to predict the target class $y$.

\subsection{Test-time interaction}
\label{subsec:test_inter}
At test time, we introduce two types of interaction to improve the prediction quality: Spatial-level interaction and Concept-level interaction.
We illustrate the test-time interaction in Fig.~\ref{fig:spatial_inter}.

\noindent \textbf{Concept-level interaction.}
Consider this scenario: While reviewing the explanations from CSR for a chest X-ray $x$, the doctor notices that the model is considering a \textit{nodule} on $x$ for a \textit{mass} in the reference image with a high similarity score.
Recognizing the incorrect comparison, the doctor informs the model that concept \textit{mass} is absent in $x$.
The model suppresses its confidence in concept \textit{mass} to calibrate its prediction.
This concept-level interaction allows doctors to directly eliminate irrelevant concepts.

\noindent\textit{Interactivity mechanism.}
When the doctor goes through $[s^{k_m}]$, she can discard any invalid concept if presented.
The model then sets $s^{k_m} = 0$ for those rejected concepts,
effectively discounting them on the prediction $\hat{y}$.
Concept-level interaction only considers the negative type of input from the doctor: \textit{concept rejection}.
Setting a suitable value for every $s_m^k$ of $c^k$ would be a overburden.
Hence, for each concept $c^k$, we only present the highest similarity score $ \max\limits_{m}\{s^{k_m}\}$ and have the doctor recognize its presence. We zero out all of the similarity scores of the rejected concept.

\noindent\textbf{Spatial-level interaction.}
As the doctor examines the image $x$,  upon spotting an important area, she can guide the model's attention with bounding boxes. 
The model then recalibrates its prediction $\hat{y}$ by prioritizing the indicated region based on this feedback.
Conversely, reviewing the activation maps $[S^{k_m}]$, the doctor also notices the model concentrating on an irrelevant region for prediction.
To counteract the Clever-Hans effect, she marks the area as spurious, prompting the model to \textit{disregard} it in its updated prediction.
Hence, the doctor only needs to direct the model \textit{where} to focus, without specifying 
\textit{what} is in that region.
This interaction scheme is practical and avoids 
the doctor 
from inadvertently becoming a teacher rather than taking advantage of the model.

\noindent\textit{Interactivity mechanism.} As we described in the user journey above, the spatial-level interaction allows two types of input from the doctor:
(i) the \textbf{positive boxes} $\{\texttt{bb}^{+}=(\texttt{x}_1,\texttt{y}_1,\texttt{x}_2,\texttt{y}_2)\}$ for the focus areas,
(ii) the \textbf{negative boxes} $\{\texttt{bb}^{-}\}$ for the regions to ignore.
To emphasize the area specified by $\{\texttt{bb}^{+}\}$ and suppress the area specified by $\{\texttt{bb}^{-}\}$, the model reweights each element $S(h,w)$ 
with an importance map $A\in\mathbb{R}^{\mathcal{H}\times\mathcal{W}}$:
\begin{align}
    A(h,w) =
    \begin{cases}
      1 & {\text{if } (h,w) \in \{\texttt{bb}^{+}\}}\\
      0 & {\text{if } (h,w) \in \{\texttt{bb}^{-}\}}\\
      \alpha & \text{otherwise}
    \end{cases}   
    ,
\end{align}
where $\alpha \in [0,1)$ is the neutral weight for the other regions.
Setting $\alpha=0$ indicates that the doctor only considers the regions $\{\texttt{bb}^{+}\}$,
while setting $\alpha\rightarrow 1$ diminishes the importance of $\{\texttt{bb}^{+}\}$.
The new set of similarity maps $[\hat{S}^{k_m}]$ is obtained by multiplying every similarity map of $[S^{k_m}]$ with the importance map $A$:
\begin{align}
    \label{eqn:imp_weight}
    [\hat{S}] = A\odot[S]=[A \odot S^{1_1}, \ldots, A \odot S^{K_M}],
\end{align}
where $\odot$ denotes the element-wise multiplication on the spatial dimension.
As CSR is designed to select the maximum element on $S^{k_m}$ as $s^{k_m}$, giving more weight to the important region ${S(h,w)}$ increases its likelihood to be selected.
On the other hand, zeroing out the spurious region nullifies its influence on the similarity vector.
Because $S$ is bounded within $[-1, 1]$ and $A$ is bounded within $[0, 1]$, we clip negative values of $S$ to condition the model on only non-negative values.
This ensures that the $\{\texttt{bb}^{+}\}$ and $\{\texttt{bb}^{-}\}$ regions are monotonically adjusted.
The new prediction is made with $\hat{S}$ by the same inference logic as described in Sec.~\ref{sec:sim_re}.

\noindent\textbf{Discussion.}
Combining both spatial and concept interaction mechanisms 
enables a flexible interaction approach.
For example, if the doctor identifies an important region, she can draw a positive box $\texttt{bb}^{+}$ over it. 
However, if the model associates the feature within $\texttt{bb}^{+}$ with a wrong concept $c^k$, placing $\texttt{bb}^{+}$ over it would amplify the error.
Because $c^k$ is absent in $x$, rejecting it while specifying $\texttt{bb}^{+}$ will instruct the model to focus on other concepts $\{c^{j \neq k}\}$ within $\texttt{bb}^{+}$ and ignore $c^k$.

\section{Related works}
\noindent\textbf{Concept-based.}
Concept Bottleneck Model (CBM) \cite{koh2020concept} predicts the concepts before using them as support for the target prediction.
Post-hoc Concept Bottleneck Models (PHCBM) \cite{yuksekgonul2023posthoc} transfer concepts from other datasets or from natural language descriptions of concepts via multimodal models like CLIP \cite{radford2021learning}.
Yang et al.\ leverages Large Language Models to generate an exhaustive list of concepts for each class \cite{yang2023language}.
These multimodal concept-based approaches~\cite{phan2024decomposing,menon2023visual,yang2023language} utilize foundation models to enhance their capabilities.
However, foundation models are generally trained on broad domains, which presents limitations when applying them directly to medical imaging.
In this work, we focus on the research direction of vision features and explore multi-modality learning in future studies.

\noindent\textbf{Part-prototypes.}
ProtoPNet \cite{NEURIPS2019_adf7ee2d} used prototypical parts to make a final classification based on positive attributions.
ProtoTree\cite{nauta2021neural} is a globally interpretable model by design that uses prototypes with decision trees. 
ProtoPool \cite{rymarczyk2022interpretable} is similar to ProtoPNet with a tweak of the shared prototypes pool and the Gumbel-softmax trick for the differentiable property.
PIP-Net\cite{nauta2023pip} is a patch-based approach with a novel regularization method for learning prototype similarity that correlates with human visual perception.
Unlike previous methods that correlate prototypes with target classes, CSR explicitly grounds prototypes on interpretable concepts, making them semantically relevant.

\noindent\textbf{Interactivity.} CBM \cite{koh2020concept} enables local interaction as human adjusts the concept logits for individual samples, while \cite{yuksekgonul2023posthoc} allows global interaction by modifying concept weights of the model.
ProtoPDebug \cite{Bontempelli2022ConceptlevelDO} and \cite{gerstenberger2023but}
focus on part-prototype interaction, refining prototypes for enhanced performance.
Besides logits-level interaction like those methods,
CSR introduces spatial-level interaction, which is particularly intuitive and beneficial in the medical imaging domain, where detailed image analysis is of significant interest to doctors.

\section{Experiments and Results}
\label{sec:result}
In this paper, we experiment on two chest X-ray datasets: TBX11K \cite{chexpert}, VinDr-CXR\cite{nguyen2022vindr}
and a skin imaging dataset ISIC\cite{codella2018skin}.
From those datasets, we define the findings annotations as the concepts and the diseases annotations as the target classes.
Details on datasets and the training hyperparameters are in the supplementary.
We qualitatively and quantitatively demonstrate the effectiveness of the concept prototypes learning framework in Fig.~\ref{fig:compare}.

\begin{table}[t]
    \centering
    \resizebox{0.75\columnwidth}{!}{%
    \begin{tblr}{
              column{3} = {c},
              column{4} = {c},
              column{5} = {c},
              cell{2}{1} = {r=12}{c},
              cell{14}{1} = {r=4}{c},
              cell{18}{1} = {r=5}{c},
              hline{1-2,14,18,24} = {-}{},
            }
            \begin{sideways}\end{sideways}          & \textbf{Method} & \textbf{F1} $\uparrow$ & \textbf{\texttt{\#}pro.}  $\downarrow$ & \textbf{\texttt{\#}exp.} $\downarrow$         \\
            \begin{sideways}TBX11K\end{sideways}    & CBM (ind.)      & 24.5            &                    & 14                        \\
                                                    & CBM (seq.)      & 60.5            &                    & 14                        \\
                                                    & CBM (joint.)    & 88.6            &                    & 14                        \\
                                                    & ProtoPNet       & 89.3            & 600                & 600                       \\
                                                    & ProtoPNet       & 89.3            & 1500               & 1500                      \\
                                                    & ProtoPNet       & \underline{94.1} & 3000            & 3000                      \\
                                                    & ProtoPool      & 90.7            & 600                & 600                       \\
                                                    & ProtoTree(H=6)  & 94.0            & 127                & 6                         \\
                                                    & ProtoTree(H=9)  & 93.7            & 511                & 9                         \\
                                                    & PIP-Net         & 94.0             & 768               & 158       \\
                                                    & CSR             & \textbf{94.4}    & 1400           & 14                        \\
                                                    & CSR (refined)   & 94.0            & 1386               & 14                        \\
            \begin{sideways}VinDr-CXR\end{sideways} & CBM (joint.)    & \underline{50.1}             &                   & 14                        \\
                                                    & ProtoPNet       & 41.4             & 1500               & 1500                      \\
                                                    & PIP-Net         & 45.1            & 768                 & 9                         \\
                                                    & CSR             & \textbf{54.6} & 1400               & 14                        \\
            \begin{sideways}ISIC\end{sideways}      & CBM (joint.)    & 45.5            &                    & 4                         \\
                                                    & ProtoPNet       & 38.5            & 600                & 600                       \\
                                                    & ProtoPool       & 31.0            & 200                & 200                       \\
                                                    & ProtoTree(H=6)  & 66.1            & 127                & 6                         \\
                                                    & PIP-Net         &\underline{69.9} & 768                & 90                        \\
                                                    & CSR             & \textbf{71.5}  & 400                & 4                         
    \end{tblr}
    }
    \caption{Diagnostic performance on the medical datasets, measured by Macro F1-score. CSR (refined) has the doctors review the concept atlas. $\texttt{\#}$pro. denotes the number of prototypes of the model. $\texttt{\#}$exp. indicates the explanation size for each prediction.}
    \label{table:result_benchmark}
\end{table}

\noindent\textbf{Diagnostic performance.}
Due to the heavy class imbalance, we use macro F1-score as the metric for our experiments.
We report the results in Table \ref{table:result_benchmark}.
It can be seen that CSR demonstrates a strong diagnostic performance compared to other baselines across all dataset.
We also benchmark the performance of CSR when we ask a doctor to refine the learned concept atlas as described in Sec.~\ref{subsec:train_inter}.

\begin{table}[]
\resizebox{\columnwidth}{!}{%
\begin{tabular}{lcccccc}
\toprule
\textbf{Method} &
  \begin{tabular}[c]{@{}c@{}}ProtoP-\\ Net\end{tabular} &
  \begin{tabular}[c]{@{}c@{}}Proto\\ Tree\end{tabular} &
  \begin{tabular}[c]{@{}c@{}}PIP-\\ Net\end{tabular} &
  CBM &
  CSR &
  \begin{tabular}[c]{@{}c@{}}CSR\\ (refined)\end{tabular} \\ \hline
\textbf{\% PG hit} &
  8.8 &
  7.8 &
  19.5 &
  55.1 &
  {\ul 60.9} &
  \textbf{79.5} \\ \bottomrule
\end{tabular}%
}
\caption{Pointing Game hit rate on TBX11K, Tuberculosis class. CBM uses the class activation map of the highest contributing product to the class Tuberculosis. ProtoTree uses a similarity map from the top prototype at level-1 of the tree. ProtoPNet, PIP-Net, and CSR use the similarity map from the prototype with the highest contribution to class Tuberculosis.}
\label{table:pg_result}
\end{table}

\begin{figure}[t]
\centering
\includegraphics[width=0.98\columnwidth]{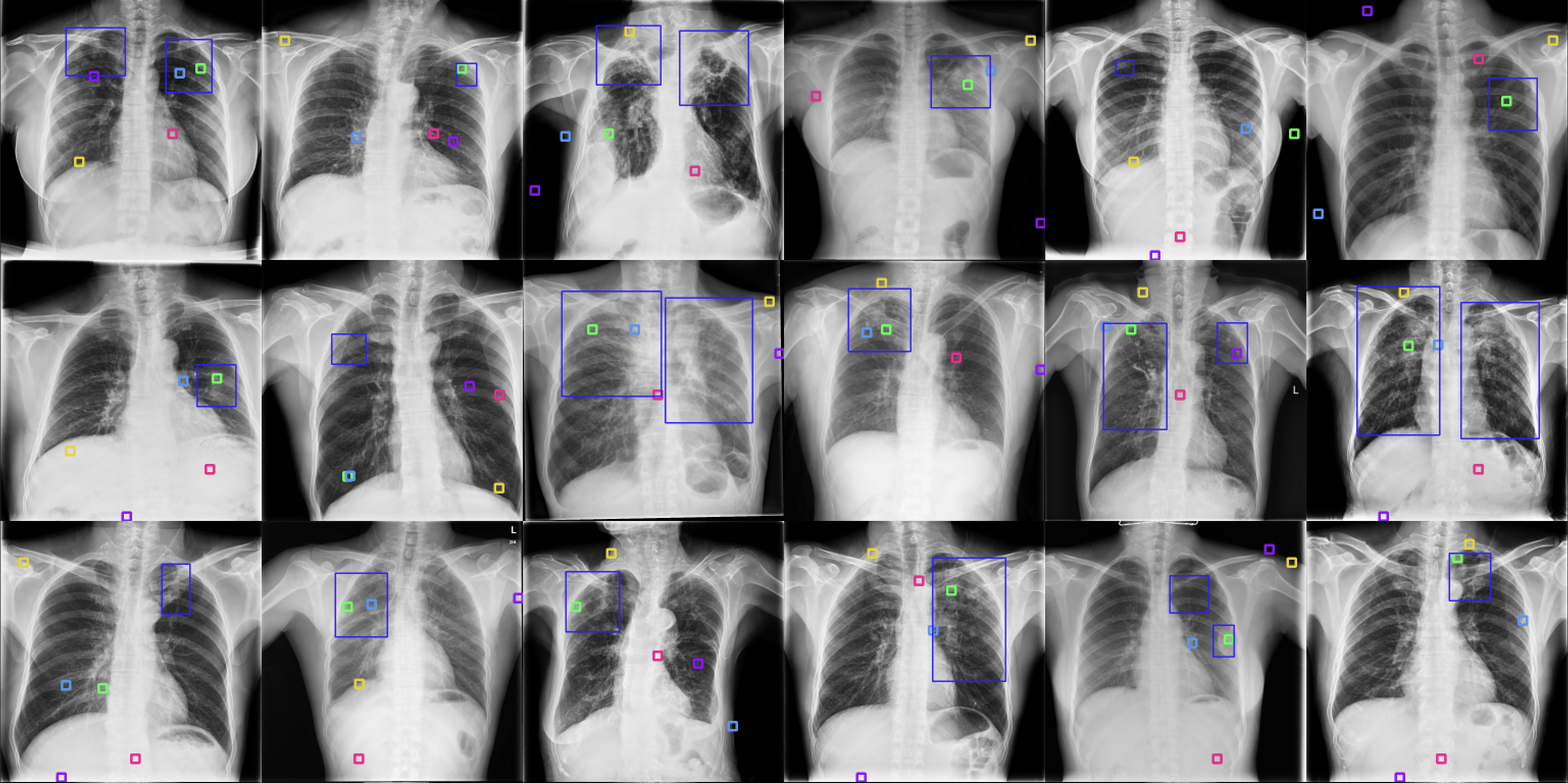}
    \caption{Qualitative result of the Pointing Game. We visualize the groundtruth bounding box and the maximum activation point of each method as the colored square. 
    $\textcolor{csr}{\square}
    \textcolor{black}{\text{:Proposed CSR, }}
    \textcolor{cbm}{\square}
    \textcolor{black}{\text{:CBM, }}
    \textcolor{pip}{\square}
    \textcolor{black}{\text{:PIP-Net, }}
    \textcolor{ppnet}{\square}
    \textcolor{black}{\text{:ProtoPNet, }}
    \textcolor{tree}{\square}
    \textcolor{black}{\text{:ProtoTree}}$}
\label{fig:quali_pg}
\end{figure}

\noindent\textbf{Explanation size.}
Following \cite{nauta2023pip}, we quantify the explanation size as the number of explanations per prediction for each method.
For part-prototype methods, explanations refer to the \textit{similar parts} of the image from the training set.
ProtoTree has the explanation size equal to its depth.
PIP-Net reduces the explanation size by regularizing the sparsity during training and only considers non-zero prototypes for each prediction.
Other part-prototype approaches have the explanation size matching the large number of prototypes.
This compromises interpretability since users must navigate through a multitude of explanations within the set for each prediction.
On the other hand, concept-based models use the \textit{concepts} as their explanations, having an explanation size equal to the number of concepts, which is typically smaller.
This choice involves an accuracy trade-off due to the reduced size of the \textit{bottleneck}. 
In contrast, the proposed CSR benefits from both strategies, featuring a small explanation size equivalent to CBM while achieving competitive accuracy akin to part-prototype methods.
To achieve this, our CSR only considers \textit{concept grounded explanations} and exclusively presents the highest similar prototype of each active concept.

\noindent\textbf{Trustworthiness.} 
\label{subsubsec:trustworthiness}
We inspect the explanations of each benchmark and demonstrate that interpretability does not always come with trustworthiness.
A trustworthy model must deliver intuitive and relevant explanations that correspond to human interpretable concepts.
To quantify such property, we leverage the \textit{pointing game} (PG) \cite{zhang2018top}  that considers a hit if the point of maximum value in the explanation lies within a segmentation ground truth.
We use the bounding box annotation from the TBX11K and calculate the PG hit rate of the explanations. 

The quantitative results are reported in Table.~\ref{table:pg_result}.
Part-prototype methods including ProtoPNet and ProtoTree are relying on clues that considerably mismatch with the semantic ground truth.
This observation aligns with \cite{nauta2023pip}, where \textit{Purity} of the prototypes is evaluated as an alignment with the part annotations in the CUB dataset \cite{wah2011caltech}, demonstrating a large semantic gap between the prototypes and human-relevant concepts.
Concept-based methods attain a more reasonable PG hit rate as the concepts are human-grounded and thus align well with the semantic ground truth.
The proposed CSR faithfully aligns with human reasoning and achieves a remarkable PG hit rate of $60.9\%$.
Refining the concept atlas further improves the PG hit rate to $79.5\%$, which hints that the better classification result of the original CSR in Table \ref{table:result_benchmark} could stem from shortcuts.
The qualitative results in Fig.~\ref{fig:quali_pg} shows that our $
    \textcolor{black}{\text{CSR:}}\textcolor{csr}{\square}$ more frequently identifies the maximum activation within the ground truth box compared to other methods.

\noindent\textbf{Interactivity benefits.}
We demonstrate the benefits of interactivity by measuring the model when obtaining interaction at test time on the TBX11K dataset and report the result in Table.~\ref{table:interaction_result}.
Our experiment involves a Medical degree (M.D.) student and a non-expert in evaluating the framework.
In the oracle setting, we use the ground truth bounding box of class Tuberculosis from the dataset as a positive spatial-level interaction.
Note that we only require human interaction when the output probability of the model is indecisive: when the difference of \text{top-1} and \text{top-2} class probability is less than $0.3$.
The neutral weight $\alpha$ is set to $0.2$.
It shows that human interactions has a positive impact on the model outcome.
The interaction from the M.D. student is more beneficial to the model than the non-expert.
We consider that the non-experts can still \textit{eliminate} irrelevant CXR regions without formal radiology expertise,
and limit the non-expert to spatial-level interactions only.
The concept-level interaction is more influential than the spatial-level interaction with better gain.

\begin{table}[]
    \centering
    \resizebox{0.75\columnwidth}{!}{%
    \begin{tblr}{lcccc}
                \hline
                           & \textbf{spatial} & \textbf{concept} & \textbf{Mac.F1} & $+\Delta\%$  \\ 
                \hline
                CSR        &                  &                  & 94.4 & 0.0    \\ 
                \hline
                Non-expert & \checkmark       &                  & 94.5 & 0.1  \\
                \hline
                Oracle     & \checkmark       &                  & 94.9 & 0.5    \\ 
                \hline
                M.D.       & \checkmark       &                  & 94.7 & 0.3 \\ 
                \hline
                M.D.       &                  & \checkmark       & \textbf{95.1} & 0.7    \\ 
                \hline
                M.D.       & \checkmark       & \checkmark       & 94.9 & 0.5    \\ 
                \hline
    \end{tblr}
    }
    \caption{Interactions performance gain. spatial: spatial-level interaction, concept: concept-level interaction. We limit the Non-export to spatial-interaction only. Oracle: using the bounding box ground truth of TB as spatial-interaction. M.D: medical student.}
    \label{table:interaction_result}
\end{table}
\section{Conclusion}
\label{sec:conclusion}
This paper introduces a novel Concept Similarity Reasoning (CSR), an interpretable and interactive model for medical image analysis with similarity reasoning.
CSR provides local explanations by presenting similar concept examples with the input images.
In comparison to part-prototype approaches, our CSR presents more concise and meaningful prototypes that align better with human-relevant concepts.
This observation implies the differentiation between interpretability and trustworthiness.
The proposed interaction mechanism employed during both training and testing effectively enhances model reliability by mitigating shortcuts and improves model accuracy through domain-specific inputs.
Our method is an attempt to encourage more trust from doctors  through a natural and effective human-AI interaction,improving both accuracy and transparency.
{
    \small
    \bibliographystyle{ieeenat_fullname}
    \bibliography{main}
}


\end{document}